\documentclass[conference]{IEEEtran}
\IEEEoverridecommandlockouts
\usepackage{cite}
\usepackage{amsmath,amssymb,amsfonts}
\usepackage{algorithmic}
\usepackage{graphicx}
\usepackage{textcomp}
\usepackage{xcolor}
\graphicspath{{Figures/}} 

\usepackage[utf8]{inputenc} 
\usepackage[T1]{fontenc}    
\usepackage{hyperref}       
\usepackage{url}            
\usepackage{booktabs}       
\usepackage{amsfonts}       
\usepackage{nicefrac}       
\usepackage{microtype}      
\usepackage{dsfont}
\usepackage{graphicx}
\usepackage{subcaption}
\usepackage{adjustbox}

\usepackage{soul}
\usepackage{footmisc}
\usepackage{url}
\usepackage{wrapfig}
\usepackage{subcaption}
\usepackage{booktabs}
\usepackage{multirow}
\usepackage{multicol}
\usepackage{tabularx}
\usepackage{adjustbox}

\usepackage{url}
\usepackage{algorithm}
\usepackage{algorithmic}
\usepackage{graphicx}
\usepackage{amssymb}
\usepackage{multicol}
\usepackage{amsmath}
\usepackage{array}

\newcolumntype{x}[1]{%
>{\centering\hspace{0pt}}p{#1}}%

\def\BibTeX{{\rm B\kern-.05em{\sc i\kern-.025em b}\kern-.08em
    T\kern-.1667em\lower.7ex\hbox{E}\kern-.125emX}}
\begin{document}

\title{
Evolutionary Multi-Armed Bandits with \\Genetic Thompson Sampling
}

\author{
\IEEEauthorblockN{Baihan Lin}
\IEEEauthorblockA{\textit{Columbia University} \\
baihan.lin@columbia.edu}
}

\maketitle

\begin{abstract}

As two popular schools of machine learning, online learning and evolutionary computations have become two important driving forces behind real-world decision making engines for applications in biomedicine, economics, and engineering fields. Although there are prior work that utilizes bandits to improve evolutionary algorithms' optimization process, it remains a field of blank on how evolutionary approach can help improve the sequential decision making tasks of online learning agents such as the multi-armed bandits. In this work, we propose the Genetic Thompson Sampling, a bandit algorithm that keeps a population of agents and update them with genetic principles such as elite selection, crossover and mutations. Empirical results in multi-armed bandit simulation environments and a practical epidemic control problem suggest that by incorporating the genetic algorithm into the bandit algorithm, our method significantly outperforms the baselines in nonstationary settings. Lastly, we introduce EvoBandit, a web-based interactive visualization to guide the readers through the entire learning process and perform lightweight evaluations on the fly. We hope to engage researchers into this growing field of research with this investigation.

\end{abstract}


\section{Introduction}

As an important practical problem in many real-world applications, online learning solves the challenge that the data is only revealed in a sequential fashion and subsequently used to update the best predictive model for unseen future reward or data corresponding to the features of the data. If we consider the many real-world scenarios of this, the reward feedback serves as the only place for the agent to efficiently learn from the available historical experience in a sequential order. The sequential decision making is a field where this setting is especially important. It concerns with an environment where the agent needs to select the best available action to take at each iteration in order to maximize the cumulative reward across a period of time. 
The key to the solution of this problem is to find an optimal trade-off between two processes in the decision making: the exploitation of the learned reward correspondence from the known actions and the exploration of unfamiliar actions. The \textit{Multi-Armed Bandits (MAB)} problem describes the mathematical formulation of this framework where each bandit arm maps to a usually fixed but always unknown reward distribution \cite{LR85,UCB}, and at each step the player picks an arm to play, gets a reward feedback and updates itself accord to the feedback. 
These online learning agents update from trials and errors based on feedback in a temporal sequence, and are widely applied to applications such as user modeling, recommendation systems, epidemic control and speaker diarization 
\cite{bouneffouf2020survey,lin2020unified,lin2020voiceid,lin2020speaker,lin2021speaker,lin2021optimal,lin2021models}.

Evolutionary computation, on the other hand, solves problem with population-based trials and errors. Usually inspired by biological evolution, these global optimization learners usually start with a pool of candidate solutions and iteratively update them by stochastically removing less favorable solutions based on some notion of fitness score and introducing incremental mutations to more favorable ones in a way that mimics the natural selection process \cite{de2016evolutionary}. Because these methods collect feedback from a pool of representations instead of merely from a temporal sequence, they usually reach a larger set of solution space, yield multiple optimal solutions, and benefit from the parallelism offered by high-performance distributed computing. Since they often provide highly optimized solutions in realistic scenarios, they are widely applied in engineering \cite{yao1999evolutionary,cartwright2004applications,srivastava2009application}.

Despite the popularity of both schools of machine learning research, there haven't been much work describing their intersection, combining the effective learning of sequential data streams based on ongoing feedback and the parallel global optimization endowed by the evolutionary computation. Most of the work in this niche field have been gravitated towards how to improve the optimization criterion of the evolutionary algorithms with bandit approaches such as \cite{belluz2015operator,liu2017bandit,lucas2018n,qiu2019enhancing}. To the best of our knowledge, this is one of the first work that combines the evolutionary computational components into the online learning problem in order to directly improve the bandit algorithms. 
We build upon the Thompson Sampling \cite{thompson1933likelihood} and propose a genetic algorithm variant, called the Genetic Thompson Sampling (GTS). We evaluate the algorithm in a series of simulation environment and demonstrate a clear advantage over existing bandit algorithms in nonstationary multi-armed bandits scenarios. Lastly, we present EvoBandit, a web-based visualization app where the users can navigate the learning process of the agent through interactive visual storytelling, play around different hyperparameters in this lightweight simulation environment, and hopefully get interested in this growing field.
\section{Problem Setting}
\label{sec:problem}

The \textit{Multi-Armed Bandit (MAB)} problem describes a sequential decision making process with reinforcement learning, where at each step the agent selects an action from a finite action set and aims to maximize the cumulative reward over time (Algorithm \ref{alg:bandit}).
There have been multiple optimal solutions proposed in either stochastic \cite{LR85,UCB} or adversarial \cite{AuerC98,AuerCFS02,BouneffoufF16} formulations. Among the formulations, the Bayesian formulation attracts attention lately \cite{chapelle2011empirical} with an algorithm called the Thompson sampling \cite{T33}. Empirical evaluation \cite{ChakrabartiKRU08} and theoretical analysis \cite{AgrawalG12} show that Thompson sampling is asymptotically optimal  for Bernoulli bandits and highly competitive for multiple complex problems. 

\begin{algorithm}[tb]
 \caption{\textbf{The Multi-Armed Bandit Problem}}
 \label{alg:bandit}
 \begin{algorithmic}[1]
 \STATE {\bfseries }\textbf{for} t = 1,2,3,$\cdots$, T \textbf{do}
\STATE {\bfseries } \quad $r(t)$ is drawn according to $\mathbb{P}_{r}$
\STATE {\bfseries }\quad  Player chooses an action $a =\pi_t(t)$
\STATE {\bfseries } \quad Feedback $r_a(t)$ for only the chosen arm is revealed.
\STATE {\bfseries } \quad Player updates its policy $\pi_t$
\STATE {\bfseries } \textbf{end for}
 \end{algorithmic}
\end{algorithm}

\section{Background}

\begin{algorithm}[tb]
 \caption{\textbf{Thompson Sampling (TS)}}
\label{alg:TS}
\begin{algorithmic}[1]
 \STATE {\bfseries }\textbf{Initialize:} $S_{a'} = 1$, $F_{a'} = 1, \forall a' \in A$.
  \STATE \textbf{For} each episode $e$ \textbf{do}
 \STATE {\bfseries } \quad Initialize state $s$
 \STATE {\bfseries } \quad \textbf{Repeat} for each step $t$ of the episode $e$:
  \STATE {\bfseries }  \quad \quad Sample $\theta_{a'} \sim Beta(S_{a'}, F_{a'}), \forall a' \in A_t$ 
 \STATE {\bfseries } \quad \quad Take action $a= \arg \max_{a'} \theta_{a'}$, and
  \STATE {\bfseries } \quad \quad Observe $r \in R_{a'}$ \\
 \STATE {\bfseries } \quad \quad $S_{a}:= S_{a}+r $ 
 \STATE {\bfseries } \quad \quad $F_{a}:= F_{a}+(1-r)$
\STATE {\bfseries } \quad \textbf{until} s is the terminal state
 \STATE {\bfseries }\textbf{End for}
 \end{algorithmic}
\end{algorithm}

\begin{algorithm}[tb]
 \caption{\textbf{Genetic Algorithm (GA)}}
\label{alg:GA}
\begin{algorithmic}[1]
 \STATE {\bfseries }\textbf{Initialize:} $M_0$ as a population of $N$ randomly generated individuals in generation 0. $\gamma$ as the selection ratio. $\mu$ as the mutation rate.
 \STATE {\bfseries } Compute fitness score $f_m$ for $m \in M_0$.
 \STATE {\bfseries } Episode $t=0$.
  \STATE {\bfseries } \textbf{Repeat} for each step $t$:
\STATE {\bfseries } \quad Create $M_{t}$, the population of generation $t$.
 \STATE {\bfseries } \quad \textbf{Selection:} Select $\gamma \times N$ members of $M_{t}$ based on fitness scores and insert into $M_{t+1}$, the generation $t+1$.
 \STATE {\bfseries } \quad \textbf{Crossover:} Create $(1-\gamma) \times N$ new members by pairing $M_{t+1}$ to produce offspring and insert them into $M_{t+1}$.
 \STATE {\bfseries } \quad \textbf{Mutation:} Mutate $\mu \times N$ members of $M_{t+1}$.
 \STATE {\bfseries } \quad t := t+1
 \STATE {\bfseries } \textbf{until} the fitness of the fittest member in $M_t$ is high enough.
\end{algorithmic}
\end{algorithm}

\subsection{Thompson Sampling}

Thompson Sampling (TS) \cite{thompson1933likelihood} describes a class of probability matching algorithm within the Bayes-optimal framework of the multi-armed bandits. It attempts to directly maximizes expected cumulative payoffs corresponding to a given prior distribution \cite{chapelle2011empirical}. As in Algorithm \ref{alg:TS}, the Thompson Sampling agent picks the arm (i.e. action) that maximizes the expected reward based on a randomly drawn belief, which in this case, is a Beta distribution. For each arm of the bandit, there are two parameters, $S$ which stands for ``success'' and $F$ which stands for ``failure''. They are initialized as 1. In each round, the agent select the action with a policy that maximizes a random variable sampled from the Beta distributions of each bandit arm. Then the agent observe a reward feedback ranging from 0 to 1, and use that to update the two parameters $S$ and $F$ by incrementing one, the other, or both, with the reward magnitude. We see that for each arm, the relative difference between the $S$ and $F$ determines the quality of the bandit arm. If the numbers of $S$ and $F$ are small (as in the early rounds of learning), the sample from the Beta distribution is very stochastic, hence promoting the exploration. When there are more evidence accumulated on certain bandit arm,  the numbers of $S$ and $F$ will increase to a large values, which make the sample from the Beta distribution more certain. 

\subsection{Genetic Algorithm}

Genetic algorithm (GA) describes a class of evolutionary algorithm that mimics the natural selection process in biology where the genomes within the population evolves by competition, selection, mutation, and crossover of genetic components \cite{mitchell1998introduction}. As in Algorithm \ref{alg:GA}, the genetic algorithm maintains an evolving population of candidate solutions. For each generation $t$, a fitness score is computed for each candidate model $m \in M_t$, and only a subset of the candidate models, deemed as elites fit enough by the fitness scores, are kept in the population of the next generation. The not-so-fit models which doesn't match the criterion are eliminated, making room for new individuals. These fitter models are then used as ``parents'', to be paired up to generate ``offspring''. These offspring are introduced into the population of the next generation until it is full. To introduce genetic diversity, the genetic algorithm then performs a number of mutations in randomly selected models in the population. This process is performed generation by generation until the fittest model in the population is fit enough for the problem.
\section{Related work}
\label{sec:related}

\subsection{Bandit algorithms and evolutionary computation}

There are multiple prior work that utilizes the bandit algorithms to improve the evolutionary computation applications. For instance, N-Tuple Bandit Evolutionary Algorithm \cite{kunanusont2017n} describes a hybrid approach that use bandit to balance the exploitation-exploration tradeoff within the search space of a large population of population with many unsampled points. This approach shows merit in game playing \cite{liu2017bandit}, automatic game improvement \cite{kunanusont2017n}, game agent optimizations \cite{lucas2018n} and modeling player experience \cite{kunanusont2018modeling}. Bandit algorithms have also been applied to adaptively select proper operator on the fly that maximizes the quality measure of the candidate solutions \cite{fialho2010analyzing}. A similar approach was later applied to more complicate problems such as multi-objective evolutionary computation \cite{li2013adaptive}. On relevant work to ours is \cite{st2014differential}, where a differential evolution algorithm is proposed to pick an agent among multiple agents whose parameters are unknown. Unlike this black box portfolio selection problem, our method directly optimizes and update the bandit agent with explicit genetic algorithm approaches.

\subsection{Multi-agent bandits}

Our usage of multiple bandit agents in a population is related to the literature of multi-agent bandits. Multi-agent networks is a class of bandit algorithms where a pool of agents shares information to one another \cite{shahrampour2017multi}. Similar to their work, our algorithm also hosts a population of agents, and makes decisions based on the majority of the agent votes. This type of multi-agent network can also be extended to the multi-agent coordination problem, where the agent can perform learning independently from one another in different iterations but still share information to one another. For example, \cite{verstraeten2020multi} explores the neighborhood structure of such networks to improve the coordination problem. If we consider the agents living in a society, the agents can cooperate with one another in a social learning way as in \cite{sankararaman2019social}, or decide to cooperate vs. defect in social dilemma situation such as Iterated Prisoners' Dilemma as in \cite{lin2020ipd}. In most cases, the cooperative multi-armed bandits have shown to be beneficial in various settings \cite{dubey2020cooperative,dubey2020kernel}.

\subsection{Mixture of experts models}

Having a population of decision making agents is also related to the mixture of experts model \cite{miller1996mixture} which uses multiple expert learning networks to divide a problem space into separate homogeneous regions to conquer individually. The output of such models are usually moderated by multiple levels of probabilistic gating functions \cite{yuksel2012twenty}. Similar to their approach, among the population of decision making agents, we only adopt the recommendation from a subset of the agents (by taking the majority vote). 

\subsection{Routing information among bandit components}

Related to the weight sharing among agents in multi-agent networks, bandit algorithms can also be designed to incorporate multiple modules and route information among them during the online learning process. For instance, \cite{lin2018contextual} proposes a contextual bandit algorithm that host multiple embeddings trained online in batches. The agent adaptively select which deep embeddings to use given a specific context and make decisions which in turn changes these embeddings in a routing way. Similarly, in bandits, information can be propagate across the modules when feedback are sparse and therefore unavailable in certain iteration, as in the semi-supervised bandit in \cite{lin2020online}.

\section{Method}
\label{sec:method}

\begin{algorithm}[tb]
 \caption{\textbf{Genetic Thompson Sampling (GTS)}}
\label{alg:GTS}
\begin{algorithmic}[1]
 \STATE {\bfseries }\textbf{Initialize:} $M$, $Q$, $A$, $R \in [0,1]$.
  \STATE {\bfseries } \quad \textbf{For} each agent $m \in M$:
      \STATE {\bfseries } \quad \quad $S^m_f = 1$, $F^m_f = 1$.
\STATE {\bfseries } \quad \quad $q\sim U(1,Q), S^m_{a'} = q$, $F^m_{a'} = q, \forall a' \in A$.
   \STATE {\bfseries } \quad \textbf{End for}
  \STATE \textbf{For} each episode $e$ \textbf{do}
 \STATE {\bfseries } \quad Initialize state $s$
 \STATE {\bfseries } \quad \textbf{Repeat} for each step $t$ of the episode $e$:
  \STATE {\bfseries } \quad \quad \textbf{For} each agent $m \in M$:
  \STATE {\bfseries }  \quad \quad \quad Sample $\theta^m_{a'} \sim Beta(S^m_{a'}, F^m_{a'}), \forall a' \in A_t$. 
 \STATE {\bfseries } \quad \quad \quad Get recommendation $a^m= \arg \max_{a'} \theta^m_{a'}$.
  \STATE {\bfseries } \quad \quad \textbf{End For}
    \STATE {\bfseries } \quad \quad Take action $a^* = Mo(a)$, and Observe $r \in R_{a^*}$. \\
  \STATE {\bfseries } \quad \quad \textbf{For} each agent $m\in \{m\in M|a^m==a^*\}$: \\
 \STATE {\bfseries } \quad \quad \quad $S^m_f:=S^m_f+r$ \\
 \STATE {\bfseries } \quad \quad \quad $S^m_f:=F^m_f+(1-r)$ \\
 \STATE {\bfseries } \quad \quad \quad $S^m_{a^*}:=S^m_{a^*}+r$ \\
 \STATE {\bfseries } \quad \quad \quad $F^m_{a^*}:=F^m_{a^*}+(1-r)$ \\
  \STATE {\bfseries } \quad \quad \textbf{End For}
  \STATE {\bfseries } \quad \quad Get fitness score $f^m \sim Beta(S^m_f, F^m_f),  \forall m \in M$. \\
  \STATE {\bfseries } \quad \quad Selection: $M^{p} = \text{selection}(M, f)$. \\
  \STATE {\bfseries } \quad \quad Crossover: $M^c = \text{crosscover}(M^{p})$. \\
  \STATE {\bfseries } \quad \quad Mutation: $M = \text{mutation}(M^{c})$. \\
\STATE {\bfseries } \quad \textbf{until} s is the terminal state
 \STATE {\bfseries }\textbf{End for}
 \end{algorithmic}
\end{algorithm}

\begin{table*}[tb]
\begin{minipage}{\linewidth}
      \caption{\textbf{MAB evaluation} with increasing population sizes (presented metric is the cumulative reward.)}
      \label{tab:pop} 
      \centering
      \resizebox{0.8\linewidth}{!}{
 \begin{tabular}{ l | c | c | c | c | c | c }
 &\multicolumn{3}{c}{Stationary bandits} \vline & \multicolumn{3}{c}{Nonstationary (NS) bandits}    \\
  & MAB-5 & MAB-10 & MAB-50  & NS MAB-5 & NS MAB-10 & NS MAB-50  \\ \hline
Random & $53.48 \pm 2.58$ & $46.36 \pm 1.93$ & $47.76 \pm 0.92$ & $66.00 \pm 6.77$ & $50.00 \pm 7.14$  & $50.00 \pm 7.14$ \\ 
TS &   $\textbf{68.40} \pm \textbf{2.40}$ &$\underline{67.94 \pm 2.55}$ & $\textbf{63.52} \pm \textbf{0.81}$  & $64.44 \pm 2.75$ & $60.20 \pm 1.21$ & $54.06 \pm 0.92$ \\ 
UCB1 & $60.62 \pm 2.12$  & $55.94 \pm 1.97$ & $48.26 \pm 0.81$ & $\underline{75.98 \pm 1.57}$ & $60.36 \pm 1.01$  & $50.54 \pm 0.67$ \\ \hline
GTS-p10 &  $66.12 \pm 2.45$ & $67.64 \pm 2.55$ & $62.64 \pm 0.74$ & $59.66 \pm 1.78$ & $69.26 \pm 1.71$  & $58.34 \pm 1.38$ \\
GTS-p25 &  $67.62 \pm 2.45$ & $66.36 \pm 2.82$ & $60.80 \pm 0.81$ & $59.08 \pm 2.76$ & $\underline{74.78 \pm 2.37}$  & $\underline{71.06 \pm 1.24}$ \\
GTS-p100 & $\underline{67.64 \pm 2.51}$ & $\textbf{68.00} \pm \textbf{2.62}$ & $\underline{61.70 \pm 0.83}$ & $\textbf{89.14} \pm \textbf{2.05}$ & $\textbf{93.84} \pm \textbf{1.01}$  & $\textbf{89.52} \pm \textbf{0.74}$ 
 \end{tabular}
}
 \end{minipage}
\end{table*}

\begin{figure*}[tb]
\centering
\includegraphics[width=\linewidth]{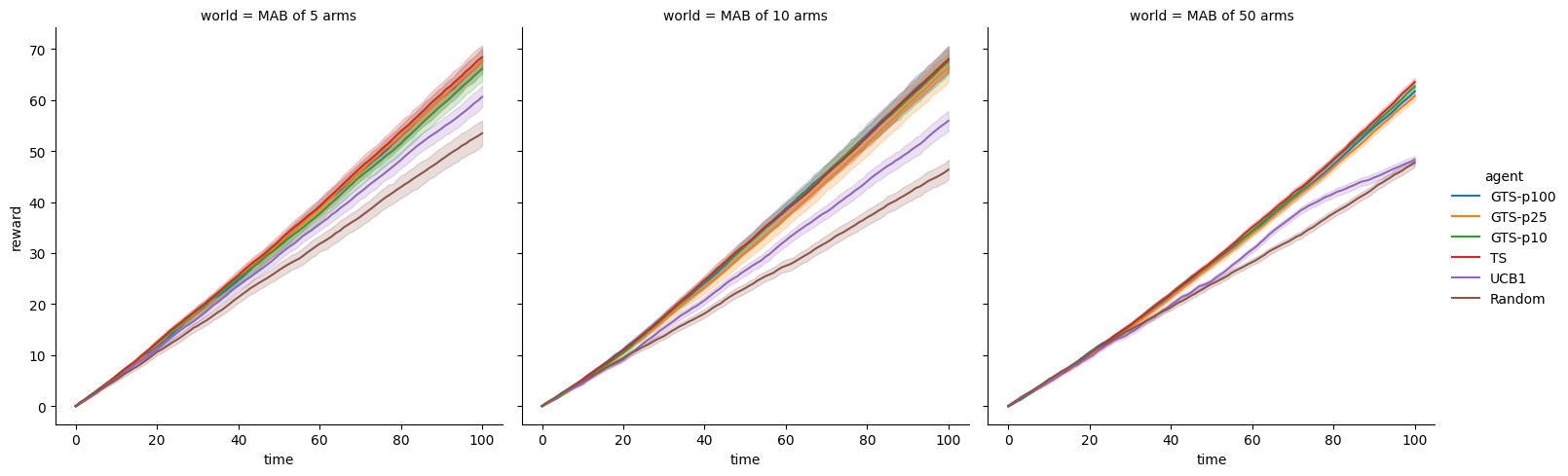}
\includegraphics[width=\linewidth]{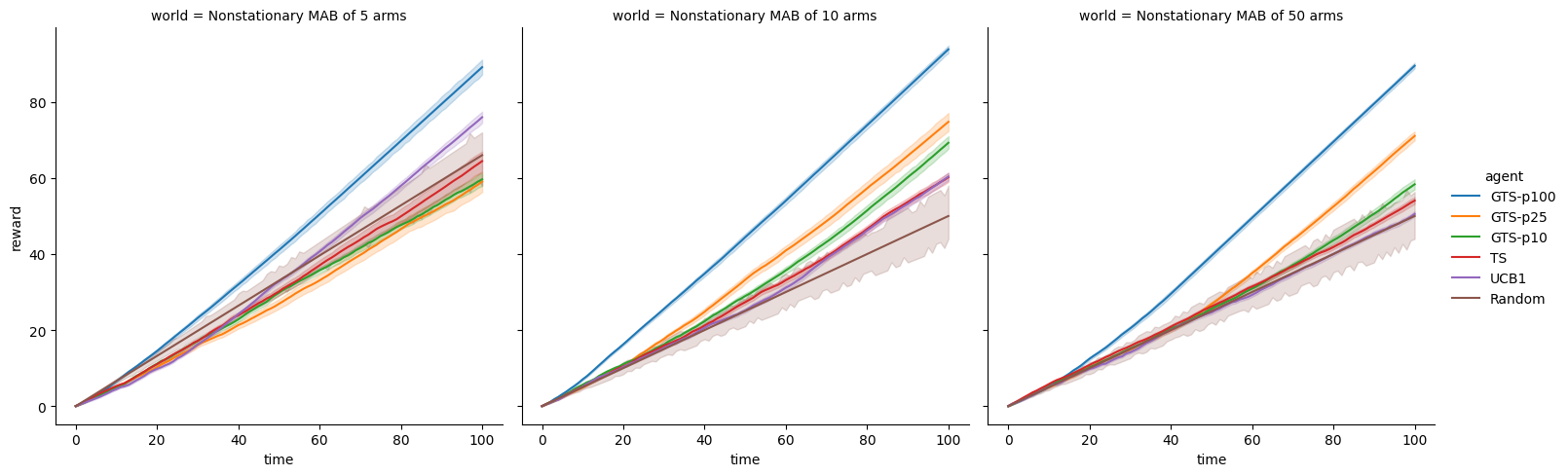}
\par\caption{\textbf{Population size.} GTS with bigger population sizes significantly outperforms baselines in nonstationary setting.}\label{fig:pop}
\end{figure*}

\begin{figure*}[tb]
\centering
\includegraphics[width=\linewidth]{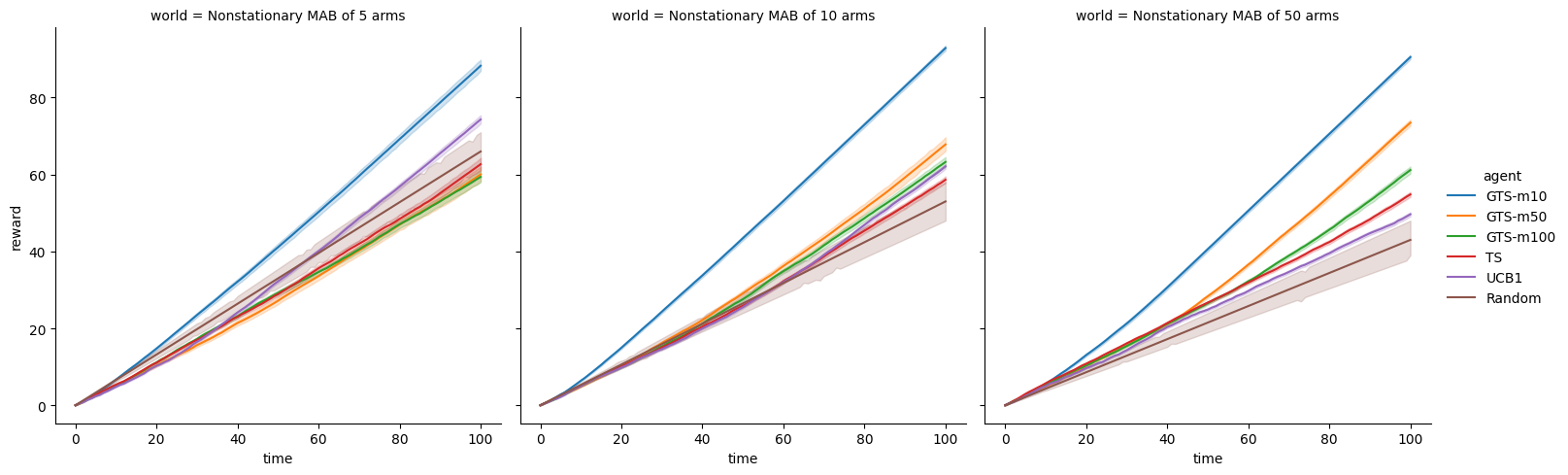}
\par\caption{\textbf{Mutation rates.} GTS is sensitive to the levels of mutation, but demonstrates consistent advantages.}\label{fig:mutation}
\end{figure*}

In Algorithm \ref{alg:GTS}, we introduce the Genetic Thompson Sampling (GTS). Here are some preliminaries: $A$ is the action space. $R$ is the reward function that assigns rewards from a range of $[0, 1]$. $M$ is a population of Thompson Sampling agents, each with their starting bandit parameters $S^m_{a'}$ and $F^m_{a'}$ for each action $a'\in A$, and their starting fitness parameters $S_f$ and $F_f$ which are both initiated as 1. $Q$ is a real number larger than 1, and the starting bandit parameters $S^m_{a'}$ and $F^m_{a'}$ are assigned by sampling from uniform distribution $U(1,Q)$. 

For each step, the pool of Thompson Sampling agents all make their recommendation of which arm to pull. Then the final action selected by the Genetic Thompson Sampling is the majority of these recommendation, i.e. the mode. Then the reward is revealed, only to update those agents whose recommendation align with the action finally chosen by the whole population. The updates have too components, one for the Thompson Sampling bandit agents, one for the fitness parameters $S_f$ and $F_f$. In another word, if the recommendation of a certain agent is adopted, and it received a positive feedback, the ``success'' rate of this agent should be higher, and vice versa (the ``failure'' rate would be higher if this is a negative feedback). Then the fitness score is stochastically computed just like the Thompson Sampling step in the bandit component, that for each agent, a sample from its Beta distribution is created, and then ranked. The rationale is that, the fitness rating can be formulated in a multi-armed bandit problem as a exploration-exploitation tradeoff. We want to select the best candidate solutions for the next step (exploitation), but we also don't want to risk not knowing what is really the best solutions as we might not have enough knowledge yet (exploration). In early round, the fitness measure would likely be less accurate than later round, and thus, should have a smaller $S_f$ and $F_f$ which makes the stochastic sampling more noisy, encouraging more exploration. In later rounds, when we have more knowledge for how fit each candidate model is, the sampling from the Beta distribution becomes more certain.

\subsection{The selection module}

There are multiple ways to determine the elites in the population to select for reproduction. In this work, we choose the most common one, by choosing the top $N$ agents based on the fitness score. The parameter we use to specify the selection is the selection ratio, and we use 0.5 (i.e. top 50\% of agents are kept as elites) throughout the evaluations ahead.

\subsection{The crossover module}

The crossover steps are as following. For each freed out spaces for new children, we randomly select two parents from the elite pool. Then, for this new agent, for each bandit arm, we randomly pick one of the two parents, and copy their bandit parameters of that arm as the bandit parameters of that arm for the child. After finishing creating the bandit parameters for this child, we initialize its fitness parameters to $S_f = 1, F_f = 1$, because we have no knowledge of how well this agent will perform just yet. (An alternative would be to use a weighted parameters that is computed as a superposition between the two parents. This would be left for future work to explore.)

\subsection{The mutation module}

The mutation models are as following. We first set a mutation rate to indicate how many mutations we want to have in this model. The higher the number, the more mutations the model will introduce. Then, for each mutation times, we randomly pick a agent from the population pool, randomly pick an arm, and then randomly assign a value from -1 to 1 to the bandit parameters of that arm.

\section{Results}
\label{sec:results}


Empirically, we evaluate the Genetic Thompson Sampling algorithm in four settings: (1) the bandits with a stationary reward function; (2) the bandits with a nonstationary reward function; (3) an ablation study of different genetic algorithm components in Genetic Thompson Sampling. We report the cumulative rewards of each agent over the learning iterations; and (4) a real-world application of the epidemic control.

\subsection{Multi-armed bandit environment}

\textbf{Simulation environments.} We first evaluate the algorithm in a simulated environment of Bernoulli multi-armed bandits. In our simulation, we randomly assign \textit{K} action arms each with a different probability of giving a reward of 1 or 0. In nonstationary environments, we change the reward distribution every $n$ rounds. We use $n=10$ throughout the evaluation.

\textbf{Baselines and variants.} We have three baselines. \textit{Random} is a random agent that picks a random action each round. Upper Confidence Bound, or \textit{UCB1}, \cite{LaiRobbins1985} and the Thompson Sampling, or \textit{TS} \cite{thompson1933likelihood}, are the two theoretically optimal solutions. We have two series of variants of the Genetic Thompson Sampling. (1) In the first evaluation, we test the effect of the population size to the agent performance. For instance, we denote the agent as \textit{GTS-p100} if the population size is 100. In this evaluation, we set the mutation rate to be 10. (2) In the second evaluation, we test the effect of the number of mutations to the agent performance. For instance, we denote the agent as \textit{GTS-m50} if the agent randomly applied 50 mutations to its population. In this evaluation, we set the population size to be 100.

\textbf{Experimental setting.} 
In our simulation, we randomly generate multiple instances of the environments and randomly initialize multiple instances of the agents. In each world instance, we let the agents make decisions for 100 steps and reveal the reward and cost at each step as their feedbacks. For all the evaluations, there are at least 50 random trials for each agent and we report their mean and standard errors. 

\textbf{Results.} 
Table \ref{tab:pop} and Figure \ref{fig:pop} summarize these results. We note that in stationary settings, the Genetic Thompson Sampling is as good as the Thompson Sampling, making the top 2 in all three scenarios. In nonstationary settings, the Genetic Thompson Sampling significantly outperforms the baselines. By varying the population size, we notice that the bigger the population size, the better. The mutation rate is more complicated, we observe that a mutation rate of 10 performs for the population size of 100 (Figure \ref{fig:mutation}.

\textbf{Ablation study.} 
We perform an ablation study on different components of the genetic updates. We denote having the crossover or not with $C+$ and $C-$, and having the mutation or not with $M+$ and $M-$. As in Table \ref{tab:ablation}, both the crossover and mutation contribute to the performance boost, but they don't account for all. Having a majority voting mechanism itself helps the learning of the Genetic Thompson Sampling. 

\begin{table}[tb]
\begin{minipage}{\linewidth}
      \caption{\textbf{Ablation study} of the components of genetic algorithm in GTS (presented metric is the cumulative reward).}
      \label{tab:ablation} 
      \centering
      \resizebox{\linewidth}{!}{
 \begin{tabular}{ l | c | c | c  }
  & NS MAB-5 & NS MAB-10 & NS MAB-50  \\ \hline
TS & $57.83 \pm 1.30$ & $57.24 \pm 1.10$ & $50.79 \pm 0.53$ \\ \hline
GTS (C-, M-) & $61.30 \pm 1.35$ & $57.72 \pm 0.89$ & $54.35 \pm 0.57$ \\
GTS (C+, M-) & $59.39 \pm 1.42$ & $63.27 \pm 1.28$ & $61.17 \pm 0.96$ \\
GTS (C-, M+) & $60.04 \pm 1.93$ & $67.83 \pm 1.78$ & $73.49 \pm 0.80$ \\
GTS (C+, M+) & $\textbf{88.29} \pm \textbf{1.54}$ & $\textbf{92.89} \pm \textbf{0.64}$ & $\textbf{90.54} \pm \textbf{0.55}$
 \end{tabular}
}
 \end{minipage}
\end{table}

\subsection{The epidemic control problem}

In this evaluation, we consider the practical problem of prescribing intervention plans during a global pandemic. This is an important real-world problem, considering how the COVID-19 has affected the lives of millions of families. The problem is as follows: say, you are a government officer, and you have at hands a series of intervention options. These options are like action dimensions: You can limit the school to two days a week, or you can close the traffic to level 3. For each action dimension, you can have different choices: say, for the traffic control dimension, you can limit it to no closure, level 1 closure (only essential traffic allowed), level 2 closure (only public transport allowed), or level 3 closure (forbid all traffics). However, each choice in each action dimension has a cost. If you close all the traffic, you may stop the spread of the virus, but the economy might crash and people might lose jobs. This is characterized by a stringency value, that evaluates how tight the government resources are. Therefore, the goal is to optimize for two objectives, to minimize the positive cases of the epidemic and to minimize the stringency level of the governmental resources. \cite{lin2021optimal} studied this question first and proposed a bandit solution for it. \cite{lin2021optimal} also introduced a simulation environment that can simulate different epidemic control scenarios and evaluate online learning algorithm.

\textbf{Simulation environments.} As described in \cite{lin2021optimal}, in the simulation environment, the user can randomly identify \textit{K} action dimension, and randomly identify \textit{N} different action levels for each action dimension. For instance, the user might design an epidemic control world where there are two action dimensions (i.e. $K=2$), traffic control and school closure; traffic control can have two levels (degree 1 and degree 2, i.e. $N^{traffic}=2$) and school closure can have three levels (all schools, all schools except universities, or all primary schools, i.e. $N^{school}=3$). The user can either consider the budget used by each intervention to be independent (each intervention approach and its value yields a fixed amount of cost regardless of other action dimensions) or combinatorial (the cost of each intervention approach and its value depends on what the action values are in other action dimensions). In real-world, the cost for each intervention approach is usually independent from other intervention dimensions. Thus, we adopt an independent assumption for these cost weights (or stringency weights as in epidemic control terms). As introduced above, the policy makers (i.e. our agents) have access to these stringency weights and thus can use them as contexts in this sequential decision making task. The epidemic control environment here can be nonstationary (which is more realistic), and it is actualized by resetting the weight matrix that maps the each action dimension and choice to the reward distributions and cost distributions. The agent receives a combined feedback of a reward and a cost. To learn from this dual signals, the agents can combine the two feedbacks into one reward given by $r^*(t) = r(t) + \frac{\lambda}{s(t)}$. 

\textbf{Baselines.} 
We evaluate four epidemic control agents. First, we have two random agents. The \textit{Random} agent randomly pick an action value in every action dimension in each decision step. The \textit{RandomFixed} agent randomly picks an action value in each action dimension at the first step, and then stick to this combinatorial intervention plan till the end. Then we have the state-of-the-art agent in this benchmark, the Contextual Combinatorial Thompson Sampling with Budget, or \textit{CCTSB}. For our agent, since our agent is only a multi-armed agent and has no contextual representation installed, we simply use it as a backbone for the combinatorial bandit problem. To be more specific, we can consider each action dimension as its own independent multi-armed bandit (an assumption that is not held in the ground truth). Then we stack these $K$ bandit agents for the $K$ action dimensions together, to form a combinatorial bandit, which we call \textit{IndComb-GTS} where ``IndComb'' stands for ``independent combinatorial'' and GTS is its backbone.

\textbf{Evaluation metrics.} 
To evaluate the problems, we report four metrics. The reward and cost are the ones recorded by the artificial environments. To better match the realistic problem of epidemic control. We post-process these two measures to create two additional metrics corresponding to real-life measures. The ``cases'' is an estimate of the number of active cases that is infected by the disease, given by an exponential function of the reward: $cases = e^{-reward^*}$, where $reward^*$ is the quantile-binned normalized reward. The ``budget'' is a quantile-binned metric of the cost. The pareto frontier would be a curve of the number of cases over the used budget.

\begin{figure}[tb]
\centering
\includegraphics[width=\linewidth]{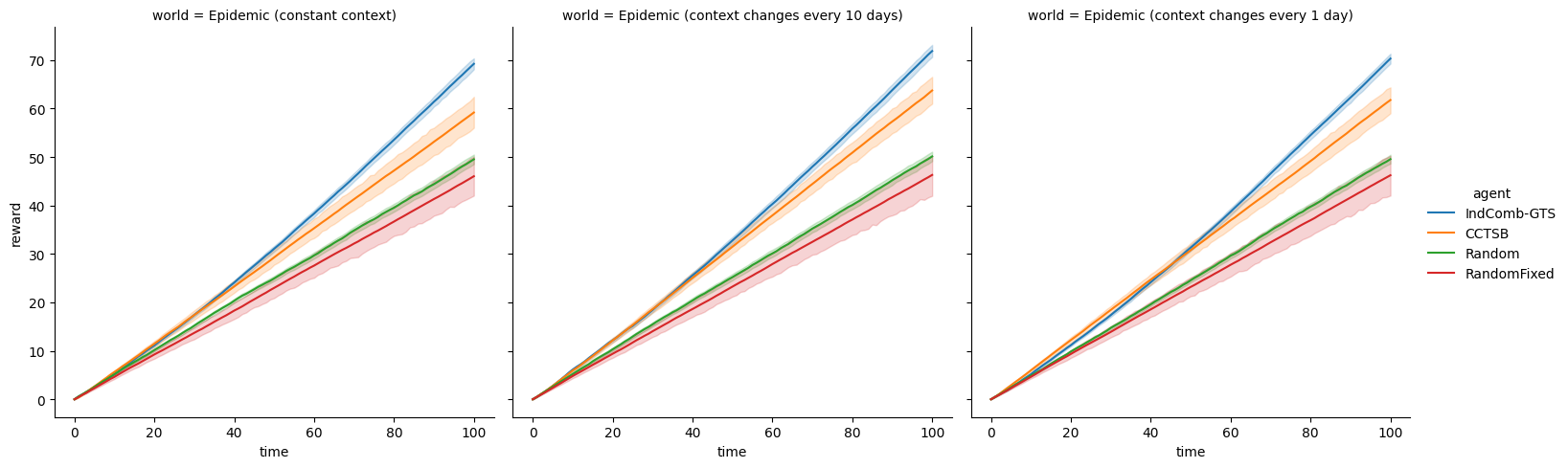}
\includegraphics[width=\linewidth]{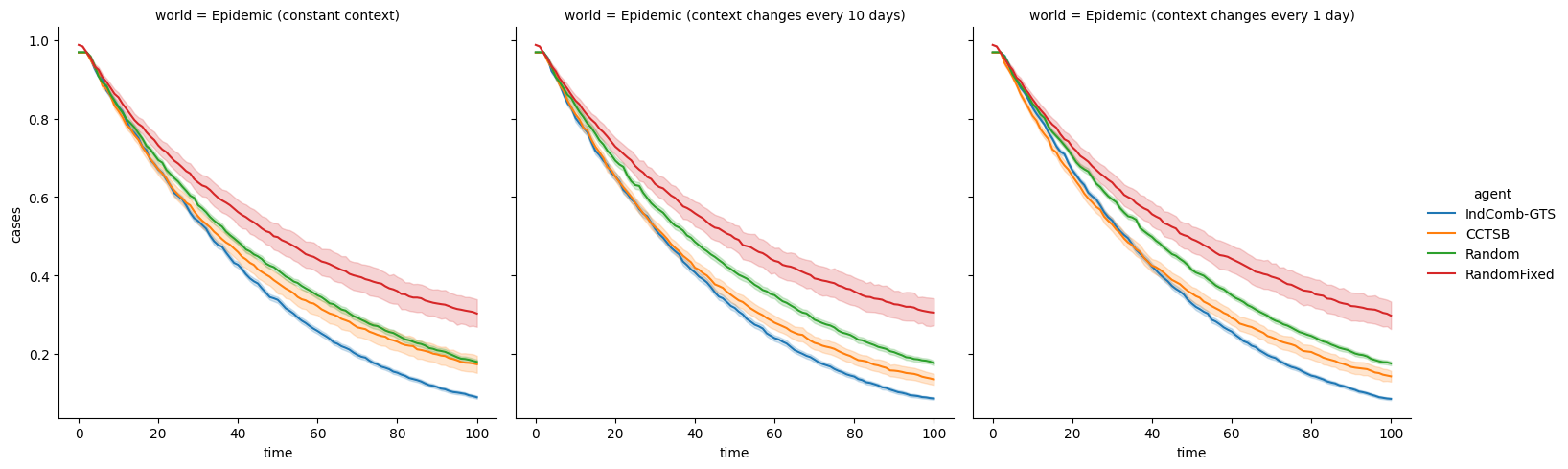}
\par\caption{Reward-driven study: GTS improves case drop.}\label{fig:w1}
\end{figure}

\textbf{Results.} 
We evaluate the agents in three scenarios. In the first scenario, we set the $\lambda$ to be 1, such that the agents are purely driven by the reward. As shown in Figure \ref{fig:w1}, comparing to the baselines, our agent IndComb-GTS significantly reduces the number of infected cases (and yields the highest rewards), which suggests that it effectively controls the epidemic spread. 

\begin{figure}[tb]
\centering
\includegraphics[width=\linewidth]{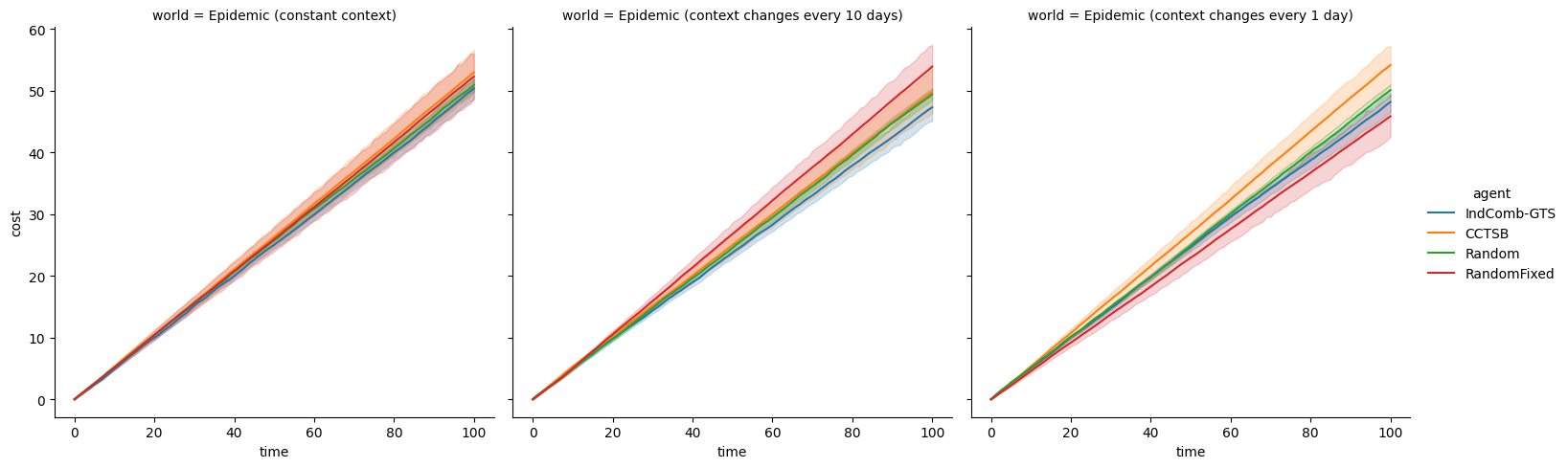}
\includegraphics[width=\linewidth]{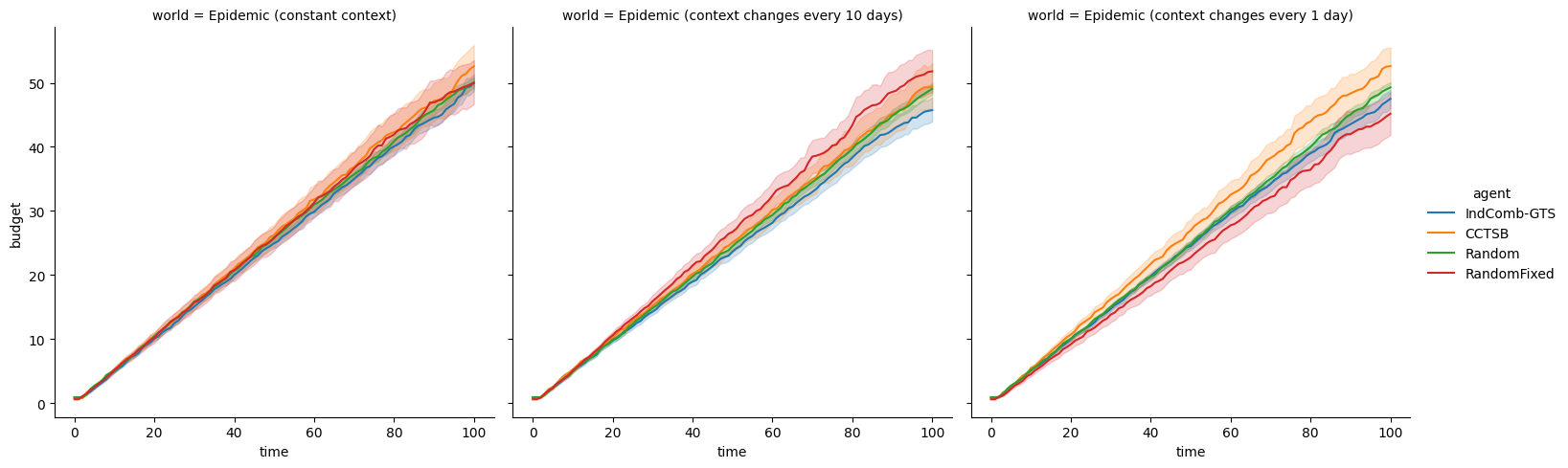}
\par\caption{Cost-driven study: GTS constrains resource stringency.}\label{fig:w0}
\end{figure}

In the second scenario, we set the $\lambda$ to be 0, such that the agents are purely driven by the cost. As shown in Figure \ref{fig:w0}, we see that it yields a similar performance with the state-of-the-art model in reducing the cost and budget. To further evaluate whether GTS can balance the tradeoff between two objectives, we perform a pareto optimal analysis for these agents.

\begin{figure}[tb]
\centering
\includegraphics[width=\linewidth]{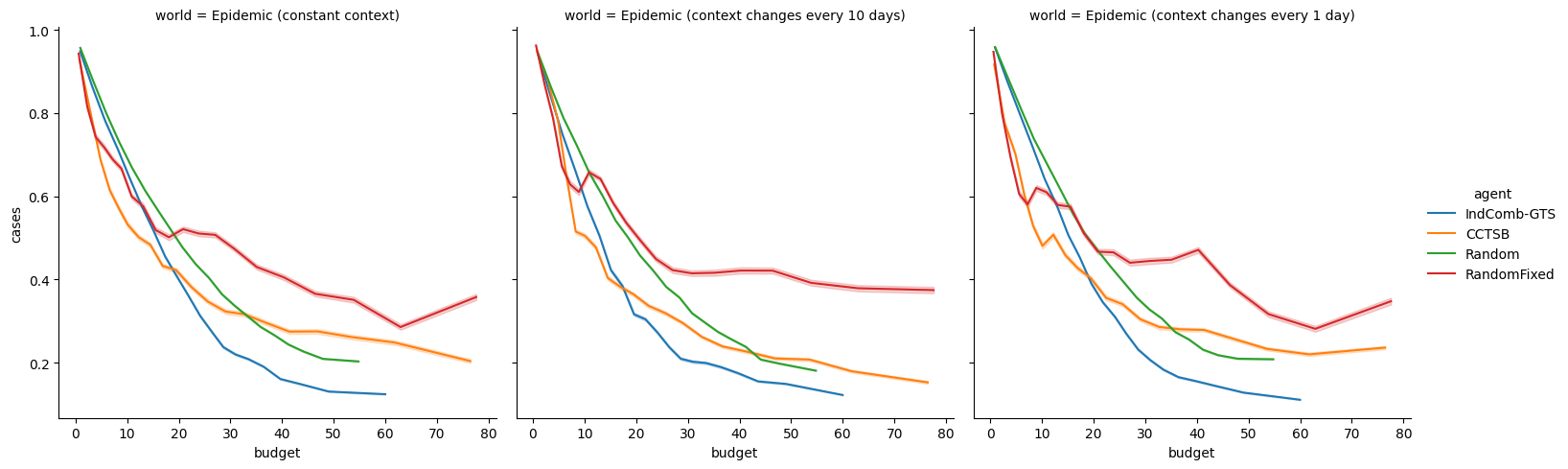}
\par\caption{Pareto frontiers of the cases vs. budget in epidemic.
}\label{fig:pareto}
\end{figure}

To obtain a pareto frontier for the agents in the epidemic simulation, we run the above evaluations with different values of $\lambda$ ranging from (0,0.25,0.5,0.75,1). Then we quantile-binned the metrics for each agent and plot out their average and standard errors. As shown in Figure \ref{fig:pareto}, our proposed algorithms yield the pareto optimal frontier, that every intervention plan extracted on its curve will minimize both the number of infected cases in a given day and the resource budget on the government.

\section{The EvoBandit Visualization System}

\begin{figure}[tb]
\centering
\includegraphics[width=\linewidth]{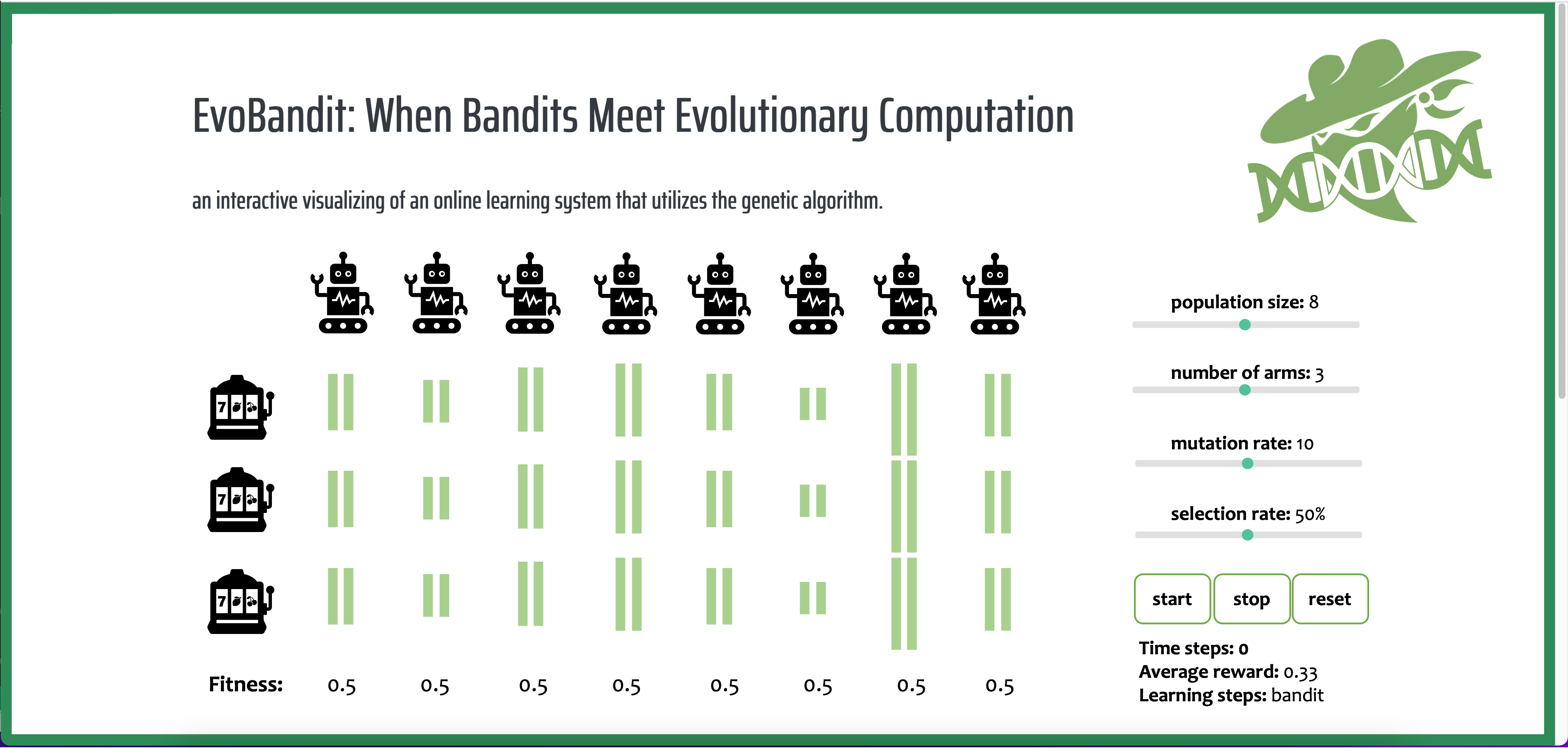}
\includegraphics[width=\linewidth]{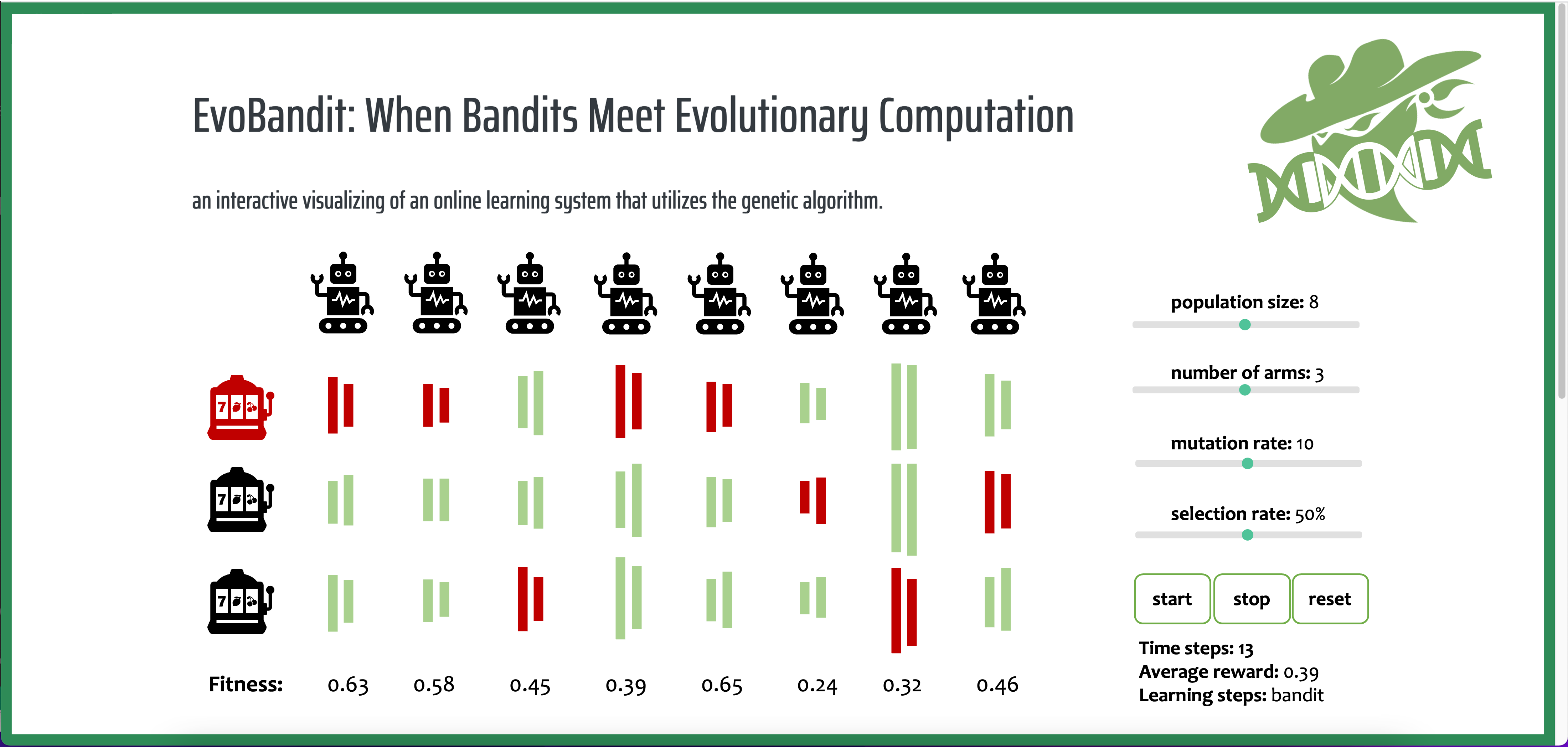}
\includegraphics[width=\linewidth]{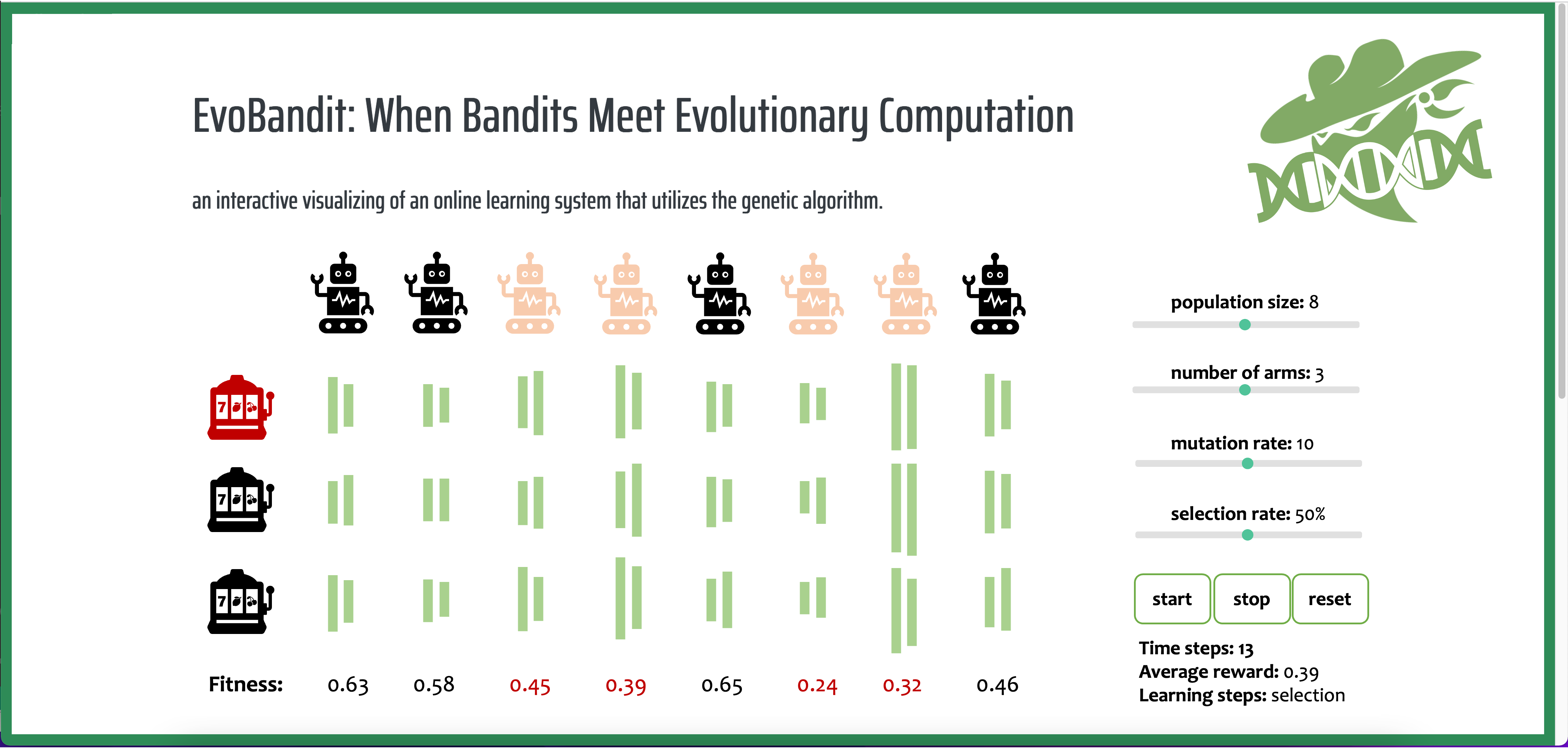}
\includegraphics[width=\linewidth]{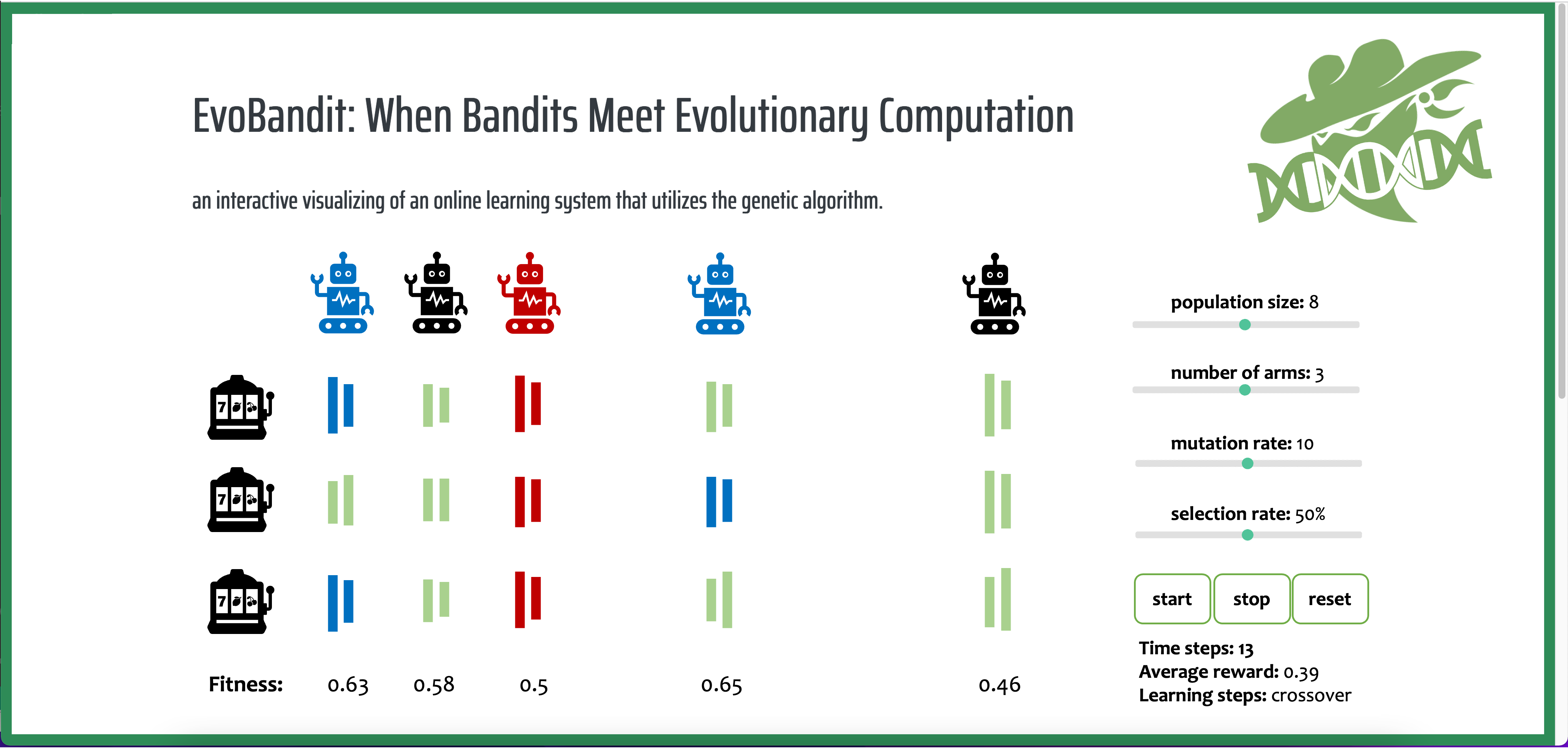}
\par\caption{\textbf{EvoBandit Visualization.} Shown here is several screenshots of the web application at different learning states.}\label{fig:demo}
\end{figure}

Here we also present EvoBandit, a web-based demonstration system to facilitate the understanding of the Genetic Thompson Sampling. As shown in the screenshots of the system (Figure \ref{fig:demo}), the user starts off by selecting a proper population size, the number of arms in this multi-armed bandit problem, and critical parameters for the genetic algorithm component of the algorithm, such as mutation rates and elite selection ratio. We limit the population size and number of arms only to a limited range, such that the users won't be distracted by too many agents and bandit and lose the main grasp of how the algorithm work. After selecting these system configurations, a grid-like representation of the evolutionary bandit agent is shown with each row corresponding to each arm of the bandts, and each column corresponding to each Thompson Sampling agent in the agent population. Each cell consists of two bars, one for $S$ and one for $F$ (as denoted in Algorithm \ref{alg:GTS}). As in the algorithm, they are each initialized with a random number (same for $S$ and $F$ for each arm within this agent), proportional to the length of the bars on this interface. As these $S$ and $F$ will increase indefinitely during the online learning process, they are rescaled properly to fit the page and only reflect their relative size among each arm and each agent. A fitness is displayed below the grid, corresponding to each agent. On the bottom right corner, there is a printout message board to guide the user along the visualization process. It contains information such as learning step, average reward, and the current stage.

When the user has input their system configurations, they can click ``start'' to begin the visualization journey. They can also pause and reset the environment any time point in the visualization. There are a few critical states. At each round, we see the agents in the population all make their recommendations, and these recommendations are labeled red. The majority of these recommendations are recorded by marking the bandit arm red. The reward is then revealed, model is then updated and the fitness scores are recomputed. Then in the selection stage, the lowest ranking agents with respect to their fitness score are eliminated, as in their pinked-out stances in the interface. During the crossover, for each new agent to fill the space, two parent agents from the elite pool are randomly selected (marked blue), and their  ``DNA'' (in this case, the parameters for each arms in this agent) are mixed by randomly selecting one parameter set from one parent for each bandit arm (marked blue as well). After all new ``children'' are introduced into the population, he mutation step (not shown in Figure \ref{fig:demo}) is simply adding a small number to the parameters of a number of randomly selected agents in the population.  

Through this interactive visualization, we believe the users can better understand the learning process of our hybrid model of bandit and genetic algorithm and get interested in pursuing this exciting growing field of research. 
\section{Conclusion}
\label{sec:conclusion}

In summary, we propose a hybrid online learning framework that combines the update principles of the genetic algorithm to a bandit algorithm. In the simulation environments of multi-armed bandits and epidemic control, this marriage between evolutionary computation and online learning algorithms appears to be a successful one in nonstationary setting. This study suggest that evoluationary components can be beneficial to the bandit learning problem and worth further investigation. Future work include extending this evolutionary bandit framework to contextual bandits, distributed systems and complex tasks. 

\clearpage
\bibliography{main}
\bibliographystyle{IEEEbib}

\end{document}